\documentclass[10pt,twocolumn,letterpaper]{article}

\usepackage{cvpr}              

\definecolor{cvprblue}{rgb}{0.21,0.49,0.74}
\usepackage[pagebackref,breaklinks,colorlinks,allcolors=cvprblue]{hyperref}
\usepackage{adjustbox}
\def\paperID{5} 
\def\confName{CVPR}
\def\confYear{2026}

\title{AttriStory: Fine-grained Attribute Realization for Visual Storytelling with Diffusion Models}

\author{
Manogna Sreenivas\quad Rohit Kumar\quad Soma Biswas\\
Indian Institute of Science, Bengaluru, India\\
{\tt\small \{manognas, krohit, somabiswas\}@iisc.ac.in}
}

\begin{document}

\twocolumn[{%
\renewcommand\twocolumn[1][]{#1}%
\maketitle
\centering
\includegraphics[width=\linewidth]{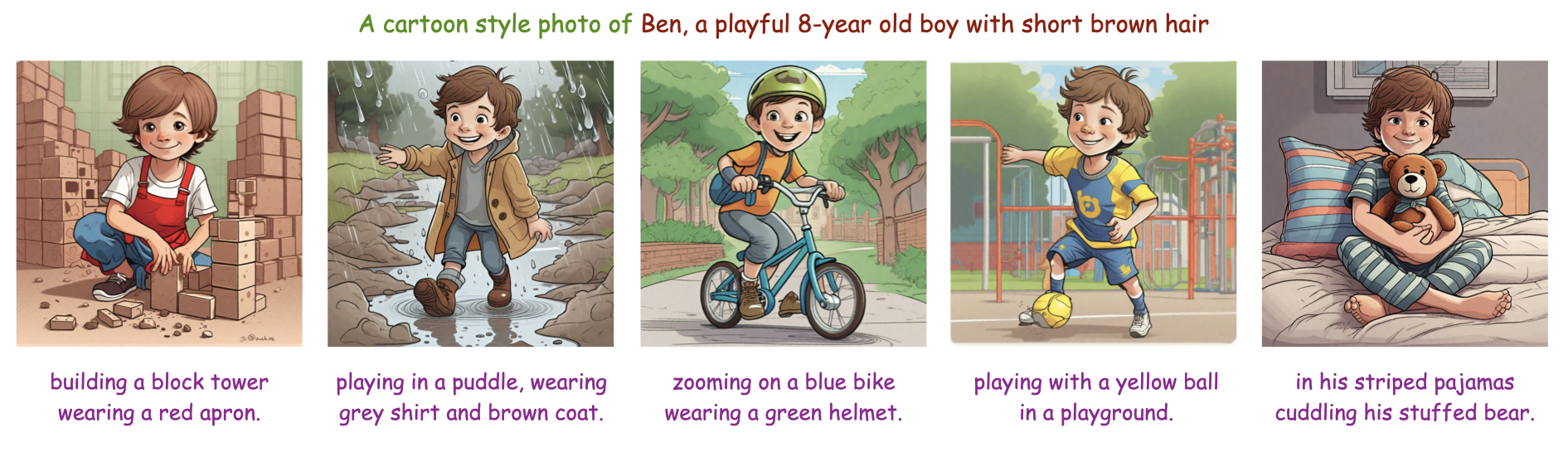}
\captionof{figure}{\textbf{Visualization of a story generated from the AttriStory benchmark. } This story of \textit{Ben}, illustrates the dual challenge in visual storytelling: maintaining character consistency across scenes, while realizing fine-grained attributes such as clothing and accessories.}
\label{fig:teaser}
\vspace{20pt}
}]

\begin{abstract}
Visual storytelling with diffusion models has made impressive strides in maintaining character consistency across narrative scenes. However, a critical gap remains: while these methods ensure a character remains consistent across scenes, they provide no systematic method to ensure if fine-grained attributes such as color and textures of clothing, accessories are faithfully rendered in the generated images. Towards this goal, we introduce AttriStory, a benchmark enabling attribute realization in visual storytelling. We curate 200 multi-scene stories across 10 distinct artistic styles using Large Language Model. Each scene is constructed with detailed attribute specifications to enable rich visual narratives. Further, to address attribute realization, we propose a plug-and-play latent optimization module that operates during early denoising steps, when the model establishes structural and semantic content. We achieve this through AttriLoss objective designed to maximize alignment between the cross-attention maps for desired attribute-object pairs while suppressing spurious associations, guiding models to localize attributes correctly. This approach operates orthogonally to existing consistency mechanisms, integrating seamlessly with current story generation pipelines without requiring architectural modifications. Our experiments demonstrate consistent improvements on incorporating AttriLoss across all baselines. This work positions attribute realization as a distinct, complementary dimension of visual storytelling, alongside character consistency, advancing the field toward fine-grained attribute-controlled story generation.

\end{abstract}    
\vspace{-10pt}
\section{Introduction}
\label{sec:intro}

Visual storytelling with diffusion models~\cite{seed-story, tale-crafter, coherent-story, story-dalle, story-gen, dragon-diffusion} has become a powerful tool for creative industries, enabling artists, animators, and designers to generate coherent story narratives with unprecedented character consistency. Recent advances~\cite{storydiffusion, consistory, 1p1s} exemplify significant progress in this direction, leveraging training-free mechanisms to preserve character identity and appearance across scenes through extended self-attention and feature sharing.

Despite these successes, we identify a critical gap in the visual storytelling pipeline: while existing methods~\cite{storydiffusion, consistory, 1p1s, characonsist} prioritize character consistency, fine-grained visual attributes remain underexplored and under-specified in both generation and evaluation. Current approaches focus primarily on basic character identity while largely neglecting details like clothing color, fabric texture, and accessories, which are elements essential for high-quality visual storytelling. This limitation becomes apparent in real-world creative workflows. In professional animation studios or editorial illustration, storytellers demand precise control over visual details. An artist might specify not merely \textit{``a boy playing with a dog"} but rather \textit{``A cartoon illustration of a boy wearing red shirt and blue sneakers playing with a brown corgi"}, where each attribute carries narrative significance. These fine-grained specifications are essential for conveying genre, details of clothing, accessories, particularly in domains like comics and animation where attribute-level fidelity directly impacts storytelling quality.

We define this challenge as the attribute realization problem: ensuring that explicitly specified, fine-grained visual attributes (e.g., \textit{``red shirt", ``blue sneakers", ``brown corgi"}) are actually realized in generated images. This problem is fundamentally orthogonal to cross-scene character consistency. While identity preservation ensures a character's recognizable features persist across scenes, attribute realization addresses whether detailed cues in the narrative text are correctly bound to visual elements in each frame. Although a model may successfully maintain character consistency, it might fail to render specified fine-grained attributes, despite explicit textual instruction. 
We address this gap, we make the following key contributions:
\begin{itemize}
    \item We introduce a benchmark named \textit{AttriStory} designed specifically to evaluate attribute realization in visual storytelling. This extends narrative scenarios with detailed attribute specifications, enabling systematic measurement of whether generated images conform to fine-grained visual descriptions. We curate 200 multi-scene stories across 10 distinct artistic styles using an LLM to generate coherent narratives with accompanying character descriptions, scene-specific actions, and positive and negative attribute-object associations (e.g., positive: \textit{[``red", ``shirt"]}; negative: \textit{[``blue", ``shirt"]}). This approach enables scalable benchmark creation by simulating realistic artist workflows where designers specify both narrative content and desired visual attributes.
    \item We propose \textit{AttriLoss}, a latent optimization objective designed to enhance attribute realization, enabling plu-and-play integration with current storytelling pipelines. This update is done during the early denoising steps when diffusion models establish structural and semantic content including colors and textures. We optimize cross-attention-based Intersection-over-Union (IoU) losses to maximize alignment between attention maps for desired attribute-object pairs while suppressing IoU between undesired attribute-object associations. Importantly, our approach complements rather than replacing existing consistency mechanisms, enabling seamless integration with recent story generation methods~\cite{consistory,storydiffusion}, requiring no modifications to their core architectures.
    \item Our experiments on integrating \textit{AttriLoss} with baseline storytelling methods and evaluation on \textit{AttriStory} benchmark demonstrate consistent and significant improvements across all evaluated baselines and artistic styles. This work positions attribute realization as a distinct, complementary dimension of visual storytelling, alongside character consistency, providing a systematic benchmark and a practical method to advance the field toward fine-grained visual narrative generation.
\end{itemize}

\section{Related Work}
\label{sec:related-work}

Visual storytelling with diffusion models sits at the intersection of consistent character generation and text-to-image synthesis. Here, we review recent advances in these areas. 

\noindent\textbf{Text-to-Image Generation using Diffusion Models.} 
Recent advances in diffusion models have revolutionized image generation~\cite{photorealistic, hierarchical, elite, dit, flux, qwen-image, animate-anyone, storymaker, free-prompt}. Denoising Diffusion Probabilistic Models (DDPM)~\cite{ddpm} established the foundation for generating high-quality images through iterative denoising. To accelerate generation, Denoising Diffusion Implicit Models (DDIM)~\cite{ddim} introduced a deterministic sampling procedure that significantly reduces inference steps while maintaining quality. Latent Diffusion Models~\cite{ldm}, exemplified by Stable Diffusion~\cite{stable-diffusion}, extended this paradigm to operate in compressed latent space, improving computational efficiency.

A key insight that enables controllable generation is the role of cross-attention mechanisms in diffusion models. Cross-attention layers bind textual tokens to spatial regions in the image through the CLIP text encoder~\cite{clip}, making them well-suited for editing. Prompt-to-Prompt~\cite{p2p} for image editing, leveraged this observation to enable training-free editing by modifying cross-attention maps during the denoising process. This work also demonstrated that early denoising steps establish coarse structure and semantic content (including colors and textures), while later steps refine fine details. 
Subsequent works~\cite{custom-diffusion, break-a-scene} have adopted similar mechanisms for personalization objectives, establishing cross-attention manipulation as a principled approach for guiding diffusion model generation.

\begin{figure*}[t]
    \centering
    \includegraphics[width=\linewidth]{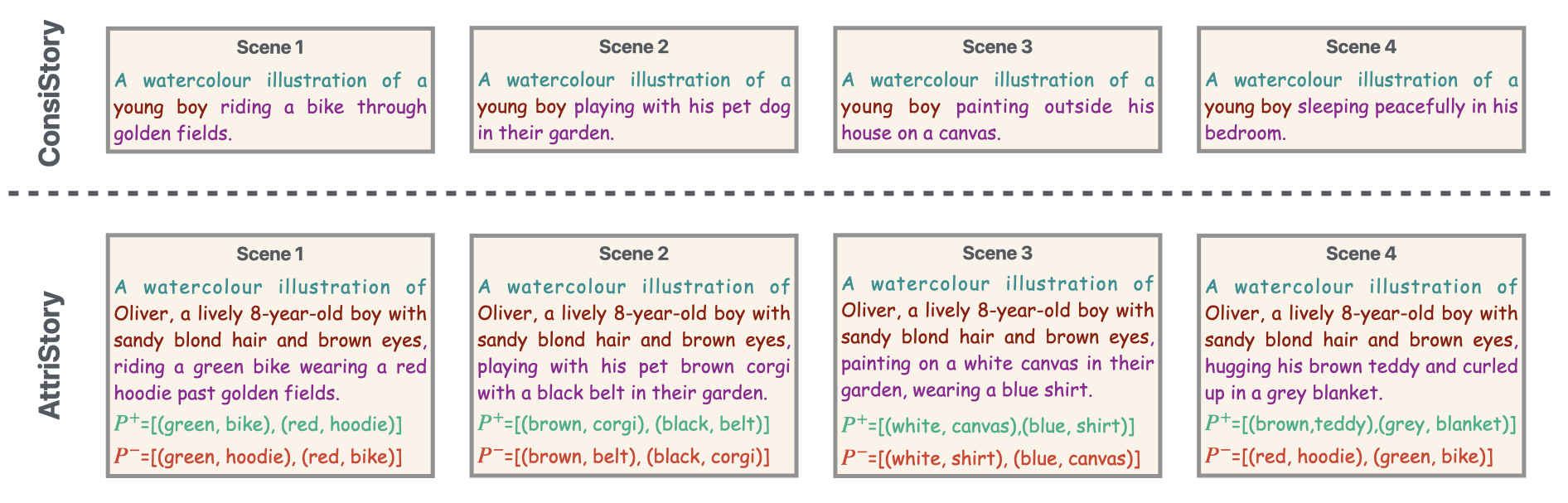}
    \caption{\textbf{Comparison of story narratives proposed in prior benchmarks vs. ours.} Existing approaches like ConsiStory (top) provide minimal 
    visual specifications, capturing only basic character identity and actions. AttriStory (bottom) enriches 
    narratives with explicit positive and negative attribute-object pairs ($P^+$ and $P^-$) for each scene, 
    enabling systematic evaluation of fine-grained attribute realization. Oliver's story demonstrates how 
    attributes like clothing color and accessories are specified and must be preserved across scenes.}
    \label{fig:consistory-vs-attristory}
\end{figure*}

\noindent\textbf{Visual Storytelling and Character Consistency.}
Character consistency across multiple scenes represents a critical challenge in visual storytelling applications. Recent research has developed two main paradigms: personalization-based approaches ~\cite{dreambooth, hyper-dreambooth, textual-inversion, break-a-scene, oracle, character-factory, instantid} and training-free self-attention based consistency mechanisms~\cite{consistory, storydiffusion, 1p1s, masactrl}.

Early personalization methods fine-tune diffusion models for specific identities. DreamBooth~\cite{dreambooth} fine-tunes the full model, while Textual Inversion~\cite{textual-inversion} learns unique embeddings. Custom Diffusion~\cite{custom-diffusion} reduces the compute overhead by fine-tuning only cross-attention projections, highlighting that cross-attention is the primary locus for identity information. IP-Adapter~\cite{ip-adapter} introduces image-based prompting through decoupled cross-attention, maintaining compatibility with other controllable generation tools. The Chosen One~\cite{chosen-one} uses iterative identity clustering to identify images with similar appearance from a set of images generated by identical prompts, extracting a consistent character representation in an automated, prompt-guided manner. PhotoMaker~\cite{photomaker} stacks identity embeddings to preserve identity while generalizing to unseen contexts, offering faster inference than test-time fine-tuning. However, these approaches require per-subject training, which is computationally expensive, hence limiting their scalability.

Recent works~\cite{consistory,storydiffusion,1p1s} has shifted focus towards training-free consistency mechanisms that are more practical for real-time storytelling applications. ConsiStory~\cite{consistory} introduces Subject-Driven Self-Attention (SDSA), which enforces character consistency by extending self-attention computations across all scenes. StoryDiffusion~\cite{storydiffusion} proposes visual memory modules that capture long-range contextual information to stabilize both character appearance and scene progression. 1Prompt1Story~\cite{1p1s} concatenates all story prompts into a single sentence with Singular-Value Reweighting (SVR) to maintain consistency, although the token length of text encoder restricts its expressiveness.

These methods represent substantial progress in preserving character identity across scenes. However, fine-grained attribute realization which is a critical need for professional storytelling, remains largely unaddressed. Our work bridges this gap by introducing both a comprehensive evaluation benchmark and a plug-and-play optimization method specifically designed for attribute realization, positioned as complementary to existing consistency approaches.

\section{The AttriStory Benchmark}
\label{sec:attristory}

\begin{figure*}[t]
    \centering
    \includegraphics[width=0.95\linewidth]{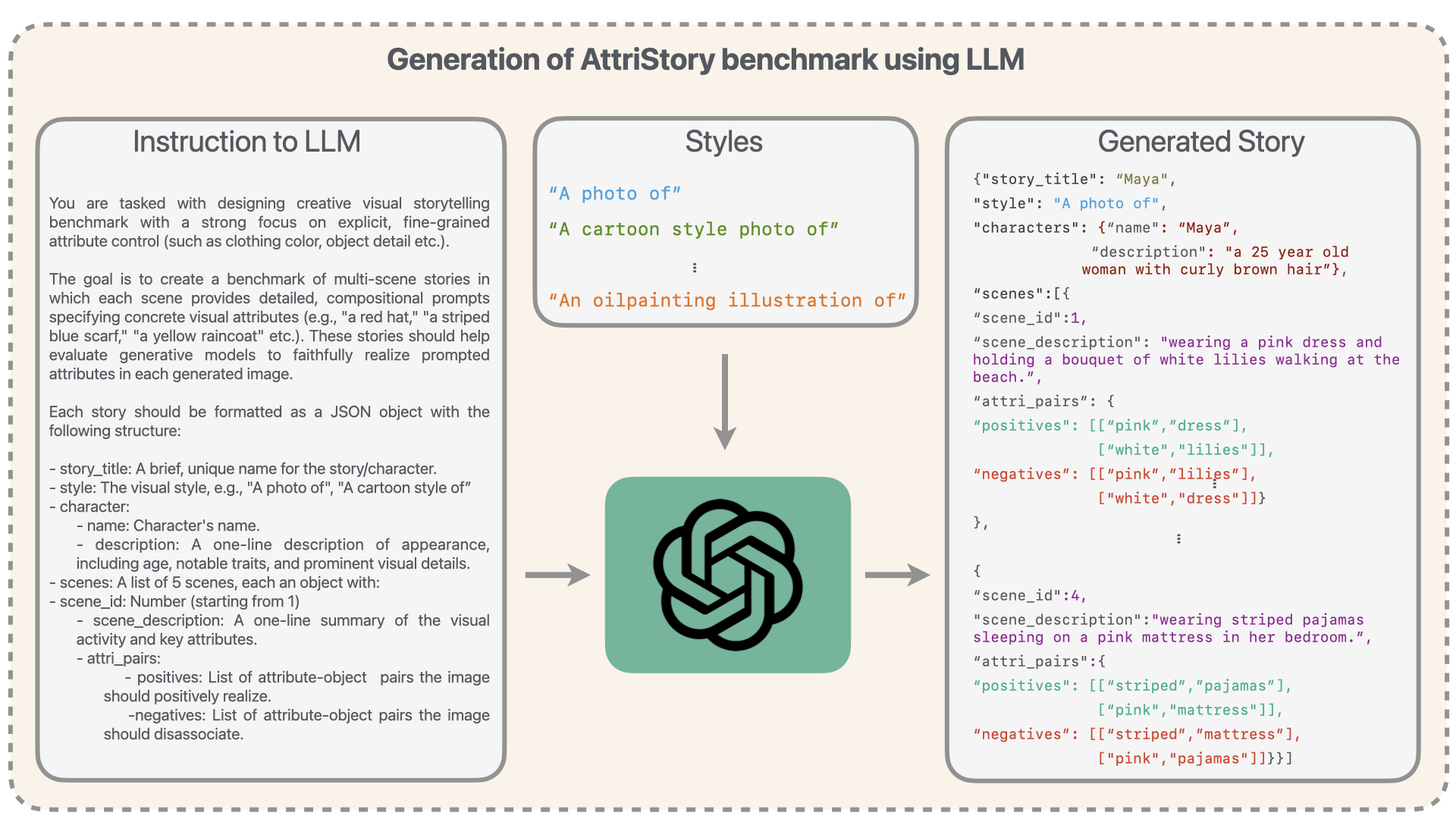}
    \caption{\textbf{LLM-driven benchmark generation. }The pipeline inputs artistic styles and structured instructions that emphasize explicit, fine-grained attribute specifications. For each story, the LLM chooses an artistic style and generates character descriptions, scene narratives, and positive ($P^+$) and negative ($P^-$) attribute-object pairs, producing structured stories enabling attribute realization.}
    \label{fig:llm-story-generation}
\end{figure*}

\subsection{Motivation}

Recent visual storytelling methods have made impressive progress in maintaining character consistency across scenes. Through attention mechanisms~\cite{storydiffusion, consistory, 1p1s}, these approaches successfully preserve character identity and appearance across narrative sequences. However, we've identified a critical gap: existing methods prioritize character consistency while largely neglecting fine-grained visual attributes specified in the narrative.

Current visual storytelling benchmarks~\cite{consistory, 1p1s} typically capture narratives through simple, minimal descriptions, such as \textit{``A photo of a young boy riding a bike through golden fields."} This level of specification is insufficient to capture how a storyboard artist would actually brief a visual team. In practice, artists communicate with rich and detailed specifications as \textit{``A watercolor illustration of Oliver, a lively 8-year-old boy with sandy blond hair and brown eyes, riding a green bike wearing a red hoodie past golden fields."} These details convey the character personality and the relationship between attributes and objects in the scene. We illustrate the difference between ConsiStory (coarse-grained) and proposed AttriStory (fine-grained) story narratives in Figure~\ref{fig:consistory-vs-attristory}, as existing benchmarks do not intend to capture this level of visual specificity.

We identify this as the \textit{attribute realization problem}: ensuring that fine-grained attributes specified in the narrative text are faithfully rendered in generated images. This is fundamentally orthogonal to character consistency. While consistency mechanisms solve the problem of \textit{the same character appearing in each scene}, attribute realization solves the problem of \textit{how that character is accessorized}, the specific visual details that carry narrative significance. 

\noindent To address this gap, we introduce \textit{AttriStory}, a benchmark specifically designed to emphasize attribute realization in visual storytelling.
This benchmark enables design of visual storytelling methods to systematically optimize for both character consistency and fine-grained visual adherence.

\begin{figure*}[h]
    \centering
    \includegraphics[width=\linewidth]{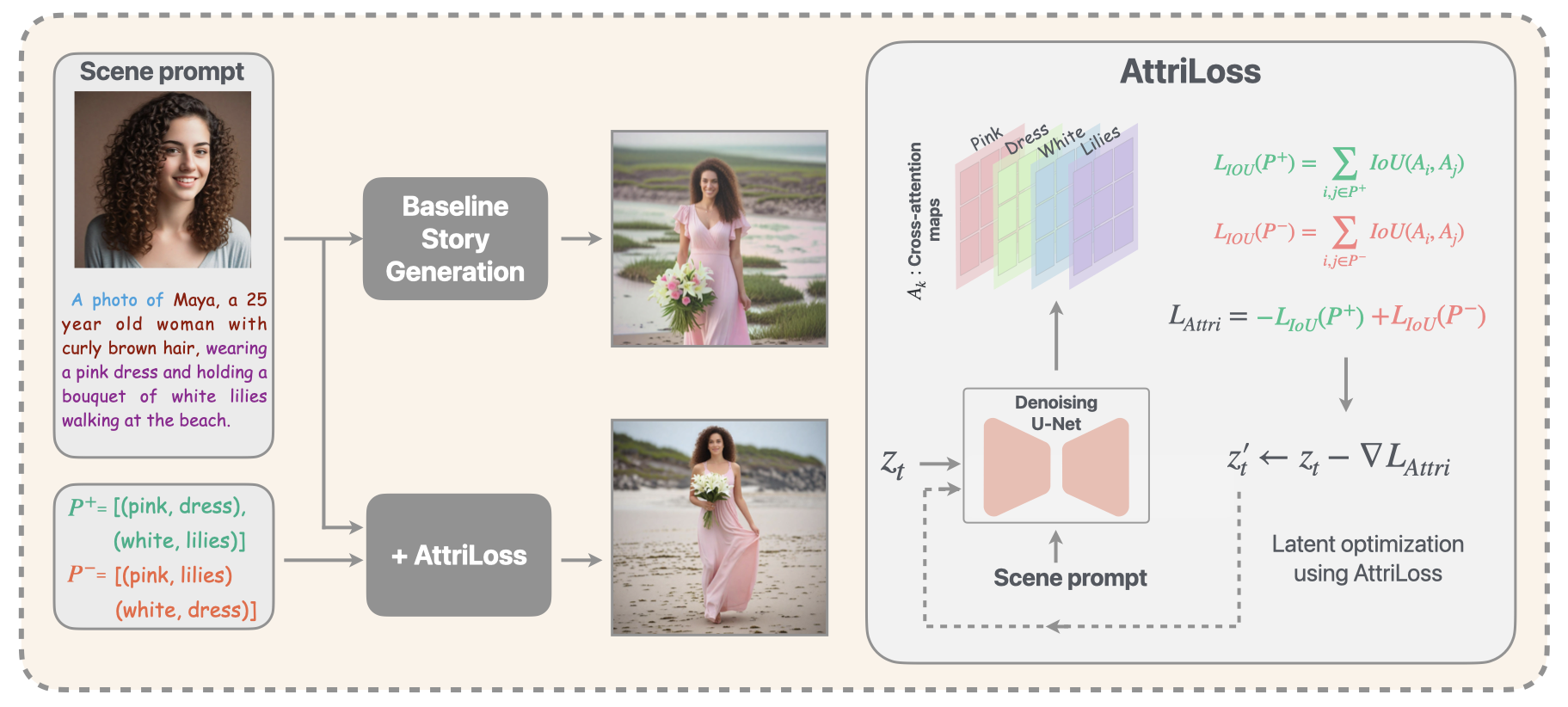}
    \caption{ {\bf AttriLoss: Targeted IoU loss on cross-attention maps. } Our method optimizes spatial overlap between attention maps of attribute-object token pairs during early denoising steps. By maximizing IoU for positive pairs (e.g., pink and dress should co-occur) and minimizing IoU for negative pairs (e.g., pink and lilies should not overlap), we guide the model to correctly localize fine-grained attributes.}
    \label{fig:method}
\end{figure*}

\subsection{LLM-Driven Benchmark Generation}

To scale benchmark creation while ensuring narrative coherence and realistic attribute specifications, we employ a large language model with structured prompts. We provide the LLM with detailed instructions specifying the story structure, artistic style, character description, and desired fine-grained attributes. As illustrated in Figure~\ref{fig:llm-story-generation}, we instruct the LLM to produce stories in the following structure:

\begin{enumerate}
    \item \textbf{Character Description:} A detailed multi-attribute character specification 
    (e.g., \textit{``Maya, a 25-year-old boy with curly brown hair"})
    
    \item \textbf{Scene Narratives:} Scene-specific text descriptions enriched with fine-grained 
    visual attributes. 
    
    \item \textbf{Positive Attribute-Object Pairs ($P^+$):} For each scene, attributes and objects that should co-occur (e.g., \textit{[``pink", ``dress"], [``white", ``lilies"]}). The attributes cover categories including colors, textures, materials etc.
    
    \item \textbf{Negative Attribute-Object Pairs ($P^-$):} Negative pairs (e.g., \textit{[``pink", ``lilies"], [``white", ``dress"]}) to prevent models from spurious associations while generation.
\end{enumerate}

\noindent We use ChatGPT~\cite{ChatGPT5} to generate the stories as illustrated in Figure~\ref{fig:llm-story-generation}. For generation, each scene prompt in a story is constructed in the following unified template: \textit{``A [artistic style] of [character description], [scene description with fine-grained attributes].''}  
The resulting stories simulate realistic creative workflows, with the goal of adhering to fine-grained specifications made in the scene prompt, given detailed information as illustrated in Figure~\ref{fig:consistory-vs-attristory}.

\subsection{Dataset Statistics and Validation}

We construct AttriStory benchmark comprising 200 multi-scene stories across 10 distinct artistic styles, each story containing 5 scenes. Each scene is annotated with 2 to 5 attribute-object pairs. The artistic styles include \textit{photo, cartoon style, 3D animation, watercolor illustration, oil painting, crayon drawing, neon punk style, pixar-style, hyperrealistic digital painting, pastel color painting}. 

To ensure high-quality story descriptions, we manually review the generated stories to ensure (1) the stories cover diverse attribute specifications in terms of color, texture, objects, (2) correctness of the attribute-object pairs, and (3) consistency between character descriptions across scene narratives. Stories failing these checks are manually revised to ensure they comply with the structure of the benchmark. This iterative process ensures that AttriStory provides a reliable benchmark to create rich visual story narratives.

\section{AttriLoss: Targeted IoU loss for Attribute-Object Grounding in Visual Storytelling}

\begin{figure}[t]
    \centering
    \includegraphics[width=\linewidth]{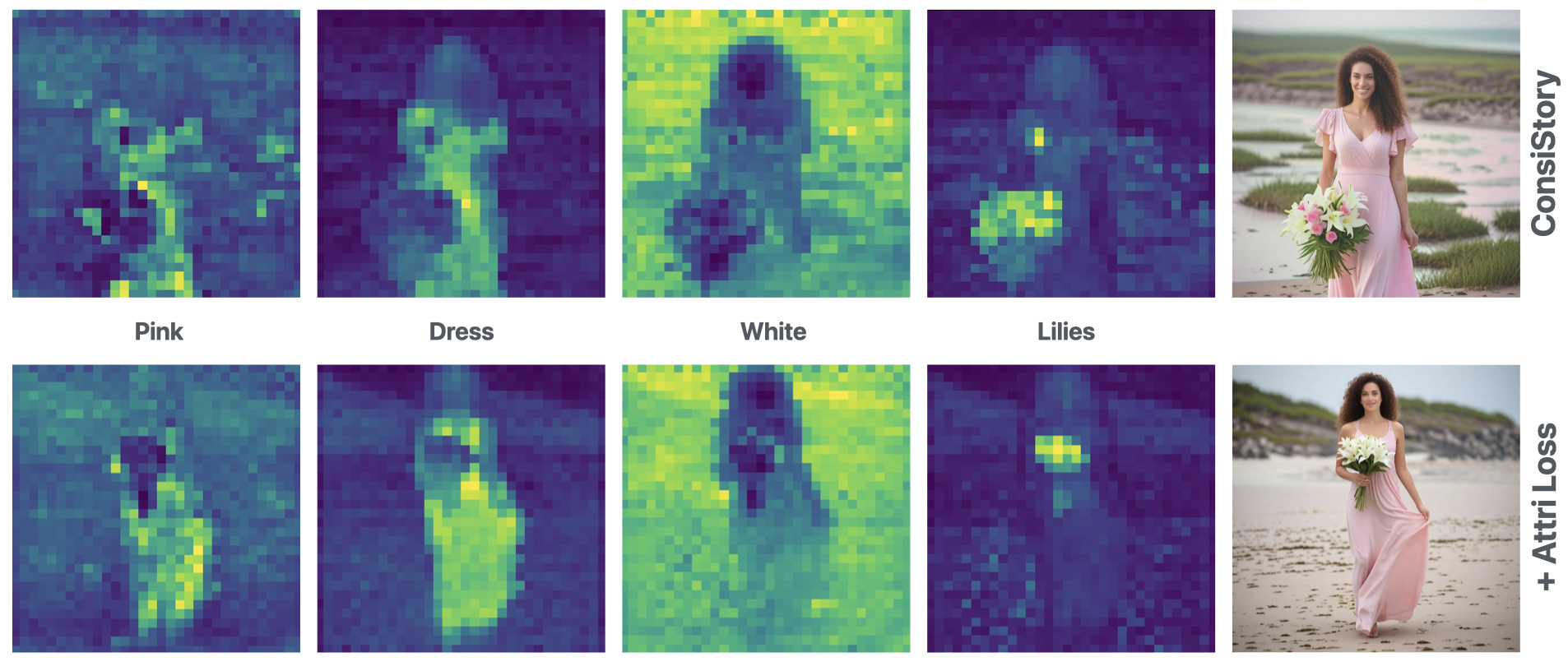}
    \caption{{\bf Attention maps of ConsiStory and with AttriLoss. } The attention maps of baseline method ConsiStory show ambiguous spatial overlaps where attribute tokens \textit{pink} and \textit{lilies} attend to the same regions resulting in the image with pink roses as well. Using AttriLoss objective with ConsiStory, the attention maps for attribute-object pairs sharpen into distinct regions (\textit{pink} and \textit{lilies} don't overlap), achieving correct spatial localization.}
    \label{fig:attention maps; before and after IoU loss}
    \vspace{-10pt}
\end{figure}

\noindent\textbf{Intuition. }
A key challenge in visual storytelling with diffusion models is imperfect alignment between prompt-specified attributes and generated visual objects. This typically arises due to the model's cross-attention mechanism: for a scene prompt like \textit{``wearing a pink dress holding a bouquet of white lilies,"} attention maps for tokens such as \textit{``pink"} and \textit{``lilies"} may overlap spatially even though \textit{``pink"} should only attend to \textit{``dress"} and \textit{``white"} to \textit{``lilies."} 
Our key observation is that the cross-attention maps corresponding to tokens provide a direct visualization of attribute-object associations learned by the model. Any ambiguous overlaps in cross-attention maps of incorrectly associated tokens can result in images that fail to ground fine-grained attributes as specified. In Figure~\ref{fig:attention maps; before and after IoU loss}, the overlap of \textit{``pink"} and \textit{``lilies"} tokens result in creating pink roses which in undesired. By explicitly measuring and manipulating the spatial overlap between attention maps of attribute-object token pairs, we can guide the diffusion process to encourage correct and discourage spurious associations.

\begin{figure*}[t]
    \centering
    \includegraphics[width=0.95\linewidth]{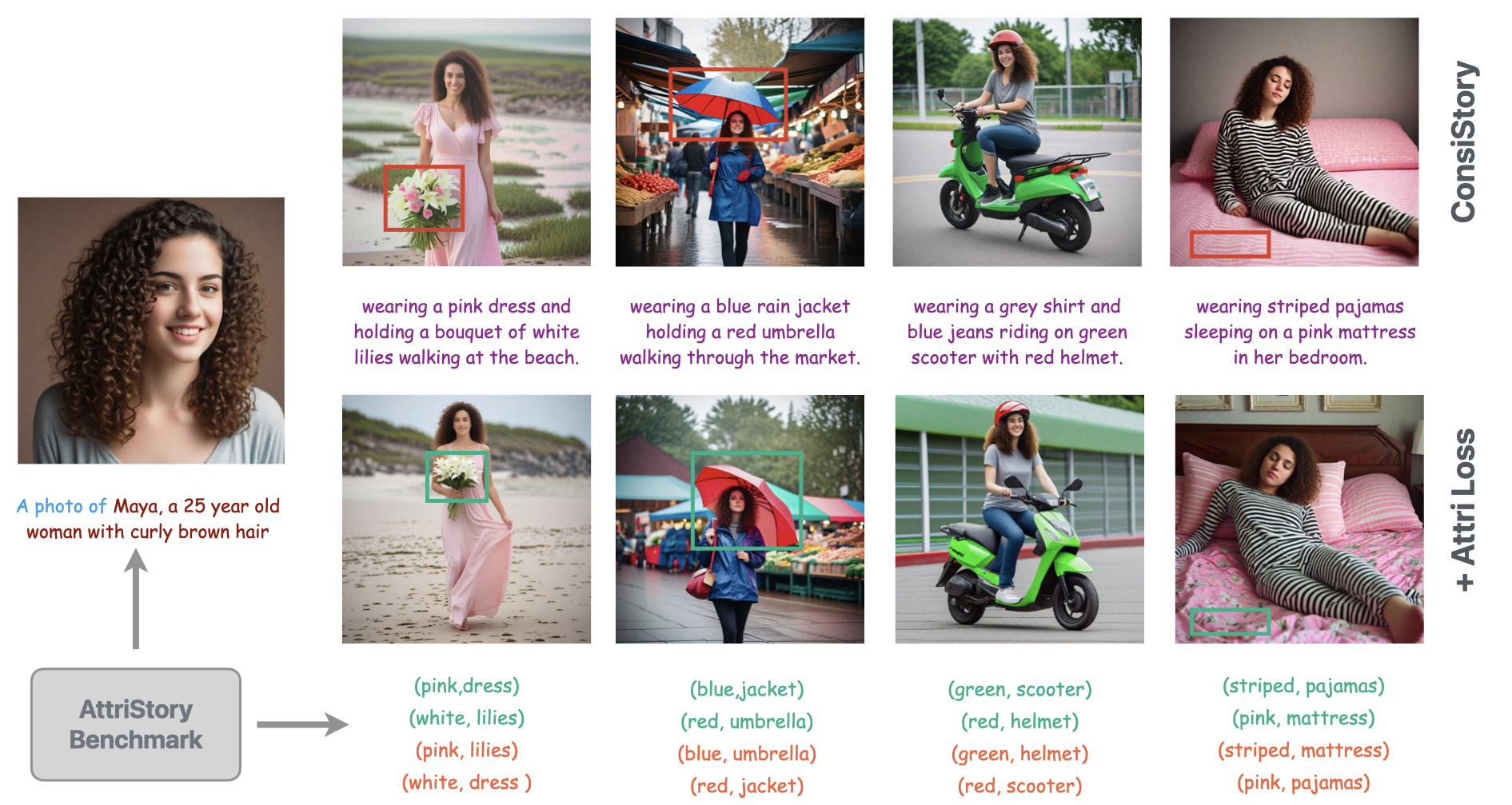}
    \caption{\textbf{Qualitative results of ConsiStory baseline with and without AttriLoss. } Using ConsiStory, the character consistency is maintained but it fails to correctly bind fine-grained attributes (e.g., pink roses are rendered with white lilies (1), umbrella is partially colored as blue instead of red (2)). With AttriLoss, attribute specifications are faithfully realized while preserving character consistency.}
    \label{fig:consistory-iou}
\end{figure*}

\noindent\textbf{AttriLoss Formulation. }
In text-conditioned diffusion models, the cross-attention layers modulate image generation, providing implicit grounding signals linking text tokens to visual features. Each token in the prompt produces a spatial map indicating the degree of attention paid to different image regions during denoising. 
We construct an objective leveraging cross-attention maps to steer generation.

\noindent Let \(P^+\) be the set of positive attribute-object token pairs that should co-occur spatially (e.g., (\textit{``pink"}, \textit{``dress"})), and \(P^-\) be the set of negative pairs that should not spatially overlap (e.g., (\textit{``pink"}, \textit{``lilies"})). For each text token \(k\), let \(A_k \in \mathbb{R}^{H \times W}\) be its spatial cross-attention map aggregated across U-Net layers while denoising.

\noindent The Intersection-over-Union (IoU) between two attention maps $A_i$ and $A_j$ is calculated as:
\begin{equation}
\mathrm{IoU}(A_i, A_j) = \frac{A_i \odot A_j}{\max(A_i + A_j - A_i \odot A_j, \epsilon)}
\end{equation}

\noindent where \(\odot\) denotes element-wise multiplication and  \(\epsilon\) is a small constant for numerical stability.

\noindent The AttriLoss objective is formulated as:
\begin{equation}
\mathcal{L}_{\mathrm{Attri}} = - \sum_{(i,j) \in P^+} \mathrm{IoU}(A_i, A_j) + \sum_{(i,j) \in P^-} \mathrm{IoU}(A_i, A_j)
\end{equation}

\noindent The latent code \(\mathbf{z}_t\) is updated via gradient descent through the proposed \(\mathcal{L}_{\mathrm{Attri}}\) objective as:
\begin{equation}
\mathbf{z}_t^{'} \leftarrow \mathbf{z}_t - \nabla_{\mathbf{z}_t} \mathcal{L}_{\mathrm{Attri}}
\end{equation}

\noindent This latent optimization steers the denoising process to maximize the IoU of positive attribute-object pairs $P^+$ and minimize the IoU of negative attribute-object pairs $P^-$. We perform this update in the early denoising timesteps (timesteps 1 to 25 out of 50 total steps) which are critical for establishing structure and semantic content~\cite{p2p}, enabling grounding of object colors and textures as specified. 

\noindent\textbf{Integration with Existing Pipelines}
Our AttriLoss objective is designed as a plug-and-play module compatible with existing storytelling pipelines that leverage self-attention mechanisms to enforce consistency in training-free manner. We integrate our module with Vanilla SDXL~\cite{stable-diffusion}, ConsiStory~\cite{consistory} and StoryDiffusion~\cite{storydiffusion}. It operates independently for each scene, refining latent code while denoising to improve attribute adherence while preserving the consistency enforced by the baseline storytelling methods. This orthogonal design allows immediate adoption without requiring architectural modifications or retraining, enabling straightforward integration into existing workflows.

\begin{figure*}[t]
    \centering
    \includegraphics[width=\linewidth]{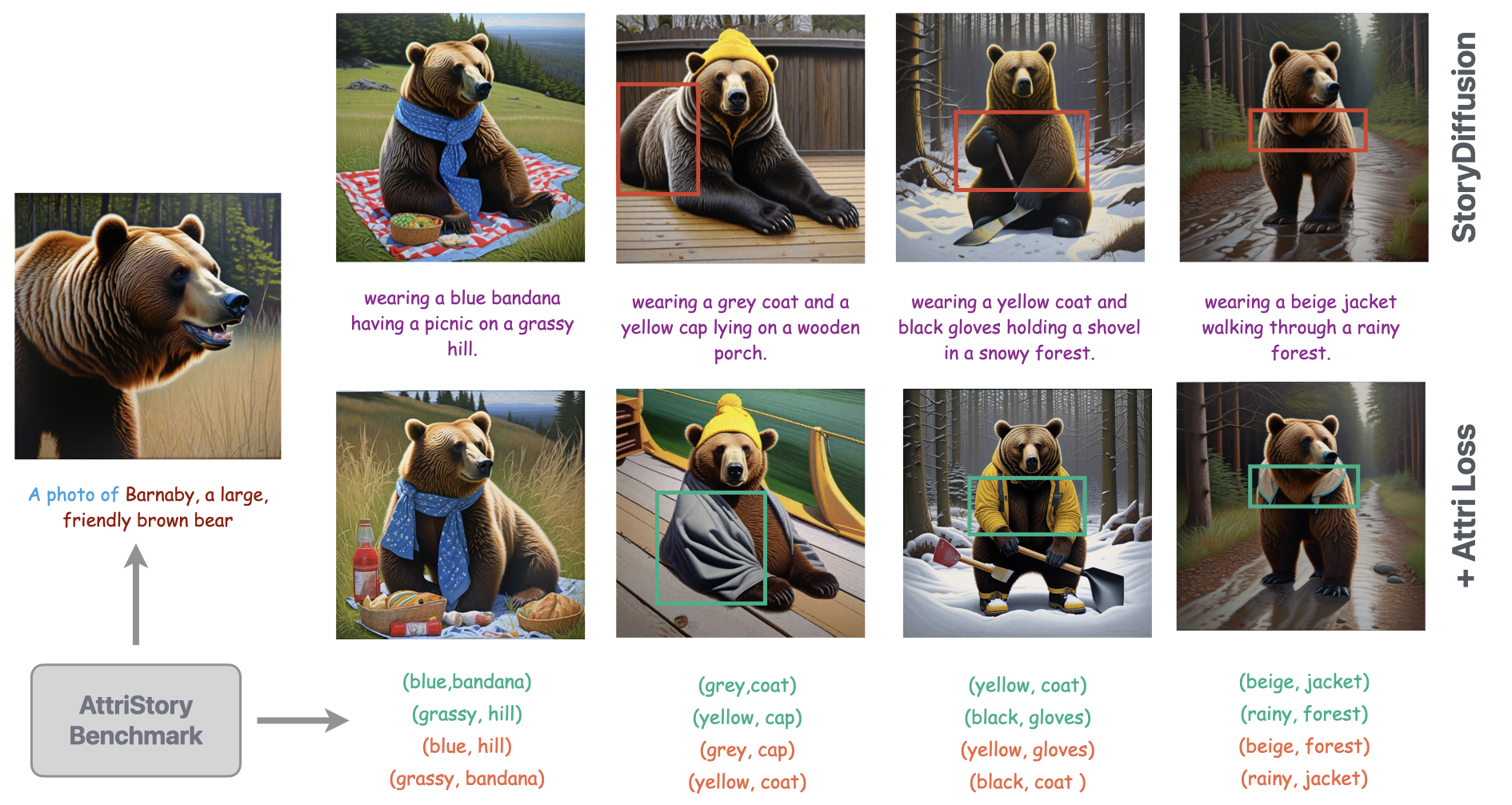}
    \caption{\textbf{Qualitative results of StoryDiffusion baseline with and without AttriLoss. } Using Consistory (top), the character consistency is maintained but fails to correctly bind fine-grained attributes (e.g., grey coat (2), yellow coat(3) and beige jacket(3) are not rendered using StoryDiffusion). With AttriLoss (bottom), attribute specifications are faithfully realized while character consistency is preserved.}
    \label{fig:storydiffusion-iou}
\end{figure*}

\section{Experiments and Results}
\label{sec:experiments}

\subsection{Implementation Details}

We evaluate three primary baselines: (1) Vanilla SDXL without consistency mechanisms, (2) StoryDiffusion with visual memory for cross-scene consistency, and (3) ConsiStory with Subject-Driven Self-Attention. We report the performance of each baseline and on integrating AttriLoss with it (denoted as ``+ AttriLoss'') in Table~\ref{tab:comparisons}. Although, we evaluate the performance of 1Prompt1story~\cite{1p1s} on our benchmark, we observe that its performance is limited in stories with detailed descriptions as ours, due its token length constraint. All experiments are conducted using Stable Diffusion XL (SDXL) as the generative model, following~\cite{consistory, storydiffusion}. AttriLoss optimizes latent codes during early denoising upto 25 time steps, using AdamW optimizer with 0.01 learning rate. All experiments are done on a single NVIDIA A6000 GPU. Evaluation is performed on the AttriStory benchmark (Section~\ref{sec:attristory}) with 200 multi-scene stories across 10 artistic styles generated using ChatGPT~\cite{ChatGPT5}.

\subsection{Evaluation Metrics}


Visual storytelling evaluation must be assessed in multiple perspectives: fine-grained attribute realization, 
cross-scene consistency, and overall visual quality. Here, we describe the four complementary metrics we employ, which provide a comprehensive evaluation of visual storytelling methods.

\noindent\textbf{Image-Text Alignment.} We use two metrics to measure whether generated images 
adhere to textual scene descriptions. CLIP Image-Text Similarity (CLIP-T)~\cite{clipscore} computes cosine similarity 
between image and text embeddings from CLIP, providing a standard baseline. However, embedding-based 
metrics compute similarity without specific focus on compositional details, limiting their 
effectiveness for fine-grained attribute prompts. VQAScore~\cite{vqa-score} reformulates alignment 
as visual question answering task: given an image and text, it converts descriptions into questions 
(e.g., \textit{``Is the dress pink?"}) and measures the probability of a ``Yes" response from a VQA model. 
This is better aligned with attribute-centric prompts proposed in our benchmark. 

\noindent\textbf{Cross-Scene Consistency.} CLIP Image-Image Similarity (CLIP-I) measures pairwise 
visual similarity between all generated images within a story, capturing whether character appearance 
and visual features persist across scenes, independent of text alignment.

\noindent\textbf{Perceptual Quality.} DreamSim~\cite{dreamsim} provides perceptual similarity 
aligned with human judgments, capturing mid-level visual properties including layout, object pose, 
color, and attribute variations. Unlike pixel-level metrics, DreamSim reflects holistic narrative 
quality.

\noindent Together, these metrics enable rigorous evaluation across attribute realization (VQAScore, CLIP-T), 
character consistency (CLIP-I), and quality measure (DreamSim).

\begin{figure*}
    \centering
    \includegraphics[width=\linewidth]{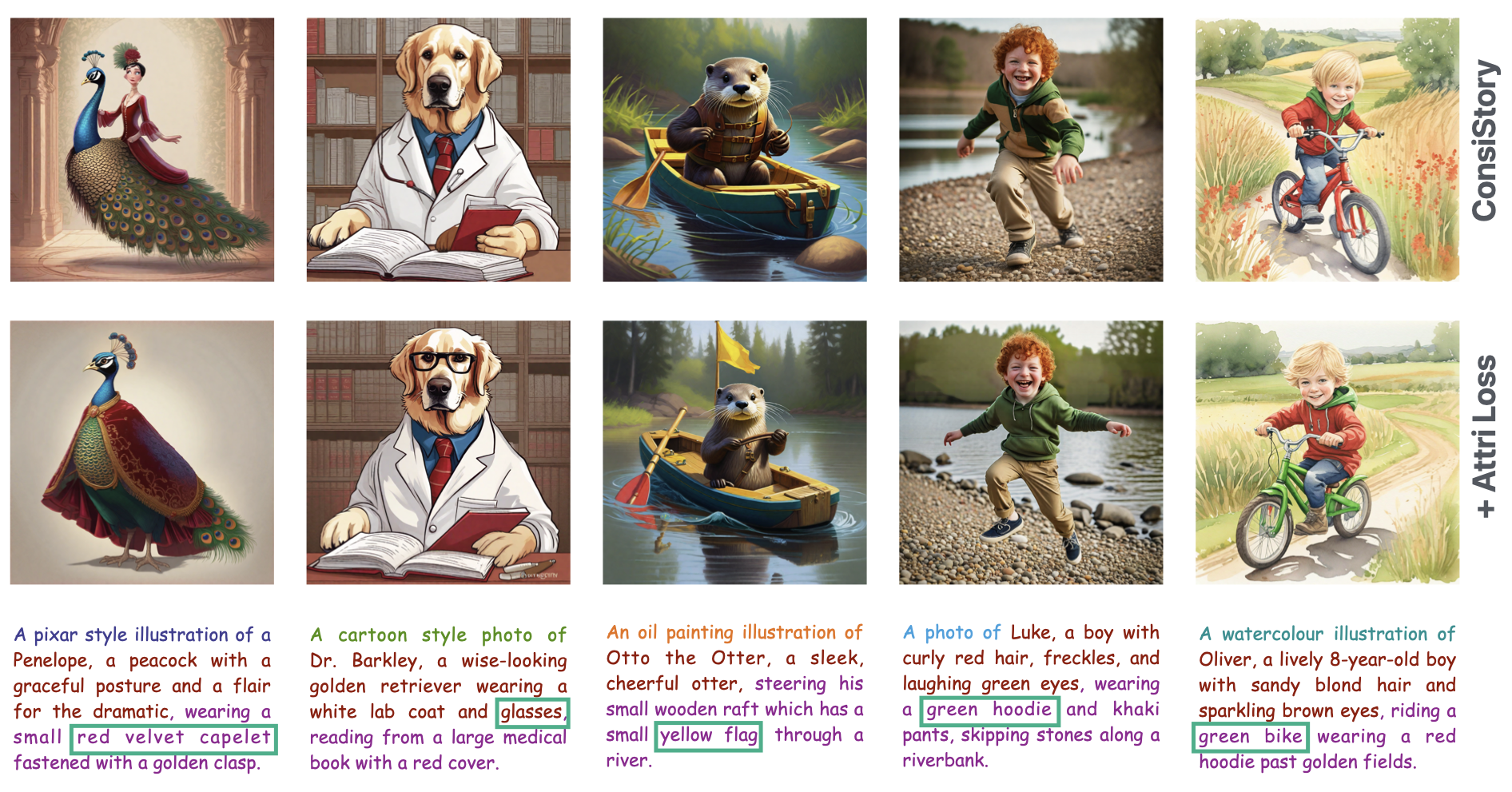}
    \caption{\textbf{Attribute realization across diverse stories} using baseline as ConsiStory (top) and with AttriLoss (bottom). Each column shows a scene in varied artistic styles (Pixar, cartoon, oil painting, photo, watercolor). AttriLoss corrects attribute-object binding failures: peacock's red velvet capelet (1), Dr. Barkley's glasses (2), yellow flag on the raft (3), Luke's green hoodie (4), Oliver's green bike (5)}
    \label{fig:additional-scenes}
    \vspace{-10pt}
\end{figure*}

\begin{table}[t]
\centering
\begin{adjustbox}{max width = \linewidth}
\begin{tabular}{lcccccccc}
\toprule
\textbf{Method}& \textbf{VQA-Score}$\uparrow$ & \textbf{CLIP-T}$\uparrow$ & \textbf{CLIP-I}$\uparrow$ & \textbf{DreamSim}$\uparrow$ \\
\midrule
1Prompt1Story~\cite{1p1s} & 0.8117 & 0.3816 & 0.8410 & 0.6929\\
\midrule
Vanilla SDXL~\cite{stable-diffusion} & 0.7957 & 0.3696 & 0.8188 & 0.6760 \\
+ AttriLoss & \textbf{0.8225} & \textbf{0.3775} & \textbf{0.8517} & \textbf{0.7170} \\
\midrule
StoryDiffusion~\cite{storydiffusion}  & 0.8363 & \textbf{0.3912}  & 0.8301 & 0.6925 \\
+ AttriLoss & \textbf{0.8636} & 0.3874  & \textbf{0.8553} & \textbf{0.7215} \\
\midrule
ConsiStory~\cite{consistory} & 0.8136 & 0.3871 & 0.8494 & 0.7326 \\
+ AttriLoss & \textbf{0.8490} & \textbf{0.3909} & \textbf{0.8667} & \textbf{0.7555} \\
\bottomrule
\end{tabular}
\end{adjustbox}
\caption{\textbf{Quantitative comparison} of evaluation metrics on integrating AttriLoss with prior visual storytelling methods.}
\label{tab:comparisons}
\vspace{-20pt}
\end{table}

\subsection{Quantitative Results}

Table~\ref{tab:comparisons} presents comprehensive quantitative comparisons across all baselines and metrics. AttriLoss consistently improves baseline performance. Significant improvement in VQAScore is associated with better fine-grained attribute realization, as it directly measures whether the specified attributes bind correctly to objects in generated images. CLIP-T shows similar or modest gains, which is expected since it is a coarse metric that does not capture compositional details, indicating that AttriLoss preserves global semantic alignment while refining attribute-level specificity. Importantly, CLIP-I scores remain strong, confirming that attribute grounding does not compromise cross-scene character consistency and validating our design that attribute realization operates orthogonally to existing consistency mechanisms. DreamSim improves across all methods, indicating that AttriLoss enhances both attribute-specific alignment and overall perceptual quality. We observe that \textit{ConsiStory + AttriLoss} achieves the best overall performance by balancing consistency and improved attribute realization.

\subsection{Qualitative Analysis}

Figure~\ref{fig:consistory-iou} demonstrates ConsiStory with and without AttriLoss on a full multi-scene narrative. The baseline preserves character consistency but fails to bind fine-grained attributes correctly:\textit{``white lilies"} are incorrectly rendered as \textit{``pink roses and white lilies"} in scene 1 and \textit{``red umbrella"} appears incorrectly colored as \textit{``red and blue"} in scene 2. With AttriLoss, attribute-object associations are properly grounded while character consistency is maintained across all scenes. Similarly, Figure~\ref{fig:storydiffusion-iou} shows StoryDiffusion with and without AttriLoss, focusing on attribute realization. Without AttriLoss, the attributes are not realized in scenes 2 to 4. With AttriLoss, compositional attributes are rendered (\textit{``grey coat", ``yellow coat", ``beige jacket"}) and consistency is maintained across scenes.

Beyond full-narrative comparisons, in Figure~\ref{fig:additional-scenes}, we presents selected scenes from different stories spanning varied visual styles: pixar illustration, cartoon, oil painting, photo, and watercolor. In each case, the baseline generates plausible characters and scenes but fails to correctly realize key attribute-object pairs: \textit{a peacock with a woman sitting on it instead of it wearing a red velvet capelet (1), Dr. Barkley's glasses are missing (2), the otter's boat missing its yellow flag (3), Luke hoodie is beige and green in color (4), Oliver's bike in red instead of green (5)}. With AttriLoss, these fine-grained attribute specifications are faithfully realized across styles, demonstrating its effectiveness.

\noindent\textbf{Limitations. } While attributes are better realized on integrating AttriLoss, there remain cases where the action described is not realized in the generated images. In Figure~\ref{fig:additional-scenes}, \textit{Dr.Barkley's face is front facing rather than reading a book (scene 2) and Otto, the otter is not actually steering his raft}, demonstrating further scope of improvement.

\section{Conclusion}

This work addresses a key gap in visual storytelling with diffusion models by focusing on fine-grained attribute realization alongside character consistency. We introduce AttriStory, a benchmark of 200 multi-scene stories with explicit fine-grained attribute-object annotations across 10 artistic styles, enabling attribute realization for visual storytelling in a systematic manner. Complementing this, we propose AttriLoss, a targeted IoU loss applied to cross-attention maps during early diffusion steps that guides models to accurately localize specified attributes. Extensive experiments demonstrate significant improvements across all baselines. By bridging identity preservation and attribute control, this work advances high-fidelity visual storytelling.



{
    \small
    \bibliographystyle{ieeenat_fullname}
    \bibliography{main}
}


\end{document}